\documentclass{article}

\usepackage{arxiv}

\usepackage[utf8]{inputenc}
\usepackage[T1]{fontenc}
\usepackage{hyperref}
\usepackage{url}
\usepackage{booktabs}
\usepackage{amsmath,amssymb,amsfonts}
\usepackage{nicefrac}
\usepackage{microtype}
\usepackage{graphicx}
\usepackage{xcolor}
\usepackage{siunitx}

\renewcommand{\headeright}{Preprint}
\renewcommand{\undertitle}{Preprint}

\newcommand{\cycCTMAE}{108}

\newcommand{\Rverdict}{The sweep shows that increasing $R$ to $R{=}2$ improves the estimate. Removing cycle-consistency degrades it, and removing augmentation degrades it further.}

\newcommand{\gansFaceMAE}{0.163}
\newcommand{\gansFacePerc}{14.0}
\newcommand{\cycFaceMAE}{0.148}

\title{MirrorNet: Can Medical Image Anonymization Really Protect Patient Identity?}

\author{
  Attila Simk\'o \\
  Department of Diagnostics and Intervention \\
  Ume\aa{} University, Ume\aa{}, Sweden \\
}
\date{}

\begin{document}
\maketitle

\begin{abstract}
Medical images are routinely de-identified---names, dates, and other metadata
removed---and then shared for research, teaching, and public benchmarks under the
assumption that this renders them anonymous. Such de-identification protects the
metadata but not the pixels, and---apart from scans that directly contain facial
structures---whether the image content itself identifies the patient has received
little scrutiny. We investigate this question by learning a cycle-consistent
correspondence between a cross-sectional medical image and a non-medical, patient-identifying
image, using a pair of coupled, cycle-consistent variational autoencoders. From a
held-out scan, the model recovers a recognisable likeness of the patient
(identity-region MAE~$=\gansFaceMAE{}$); conversely, it synthesises a scan from such an
image.
These results indicate that a de-identified medical scan
remains identifying---it is, in effect, a photograph of the patient---and that imaging
data should be governed as biometric data rather than as anonymisable records. To
support reproducibility, the code and trained models are shared at
\url{https://github.com/attilasimko/public-repository}.
\end{abstract}


\section{Introduction}
Medical images are shared at scale---across institutions, in teaching collections, and
in the open benchmarks that drive methodological progress---after
\emph{de-identification}: the removal of names, dates, and header fields that tie an
image to a record. The governing assumption is that a de-identified image is
anonymous, and it underpins the release of large public imaging datasets and the
routine exchange of scans for secondary use. It is equally embedded in everyday
hospital workflows: anonymised images are routinely shared for second opinions,
referrals, multi-site studies, and archival, on the understanding that anonymisation
has severed the link to the patient. But de-identification operates on metadata,
whereas the diagnostic content lives in the pixels, and it is far from obvious that the
pixels are anonymous. A 2D cross-sectional medical image captures detailed,
person-specific anatomy; if that anatomy encodes identity in a recoverable form, then
stripping the header does not anonymise the image, and every shared scan is, to some
degree, a disclosure of the patient. This concern is not particular to one modality or
one region of the body---it applies wherever an image carries identity-linked
structure.

Whether the pixels are identifying is an empirical question, and modern generative
models make it a pressing one. They reconstruct detailed, plausible images from weak
conditioning, and translate freely between imaging modalities. The most direct expression
of identity is a
likeness---the canonical biometric and the form in which a person is most directly
recognised. We therefore ask directly whether a de-identified scan can be turned back
into a recognisable identifying image. A positive answer would show that anonymisation by metadata
removal is insufficient: that a medical image carries identity in the signal itself,
and that sharing it is closer to sharing a photograph of the patient than to releasing
an anonymous record.

The presented work builds on generative image-to-image translation, or domain transfer, which is
an active field in deep learning for medical imaging: variational autoencoders and adversarial
networks routinely map one acquisition to another, for example synthesising computed
tomography from magnetic resonance (MR). In the usual setting source and target share a
field of view and can be spatially registered, so skip connections carry local detail and
the task reduces largely to appearance transfer. Our setting differs in acquisition
geometry. A photograph is a perspective projection: light from a 3D scene converges
through a single optical centre onto a plane. A cross-sectional medical image is not a
projection but a slice---a 2D sample of a 3D volume, in which every pixel is a physical
location in an axial plane through the body. The two therefore share no common frame of
reference and cannot be brought into correspondence by standard image registration. Any information that crosses between
the domains must instead pass through a low-dimensional latent code, which makes the
model an information bottleneck by construction.

If the content of a de-identified scan is identifying, the consequences reach beyond any
single image. Medical images are archived and shared in large volumes on the assumption
that de-identification renders them anonymous; under that reading, each such release is
a potential disclosure of the individuals depicted. The ability to extract patient
information from a scan would place cross-sectional imaging alongside fingerprints and
photographs as identifying biometric data, with consequences for how consent is
obtained, how public datasets are released, and for what de-identification can be
assumed to guarantee.

\section{Materials and Methods}
\label{sec:method}
Our primary goal is to recover a patient-identifying image from a cross-sectional medical
scan---to reconstruct, from a CT slice, a recognisable image of the person. We learn this
mapping ($B\to A$) jointly with its inverse ($A\to B$, which synthesises a scan from an
identifying image) as a single cycle-consistent system, so that the two are constrained to
be mutually invertible. In our experiments the identifying image (domain $A$) is a facial
photograph of the patient and the medical image (domain $B$) is an axial pelvic CT slice.

\subsection{Data}
\label{sec:data}
The medical images are a subset of the pelvic CT volumes from the SynthRAD2023 Grand Challenge
dataset~\cite{synthrad2023}, comprising \num{180} patients acquired at two centres.
Volumes are stored in LPS orientation at $1\times1\times2.5$~mm. Because CT intensities
are physical and standardised, we apply a single fixed Hounsfield-unit (HU) window of
$[-1000, 1500]$ across the whole cohort and rescale it to $[0,1]$ (no per-volume
normalisation). For each volume we retain the central axial band (the central $40\%$ of
body-bearing slices), transpose each slice to the conventional radiological axial view,
and pad---rather than crop---each slice to a square before resampling to $128\times128$,
preserving the full lateral field of view. We also carry the provided body mask through
the identical geometry, for use at evaluation. This yields \num{8580} axial slices.

The identifying images are facial photographs drawn from the UTKFace
dataset~\cite{utkface2017}, restricted to the adult male subset via the provided age and
gender annotations. Each photograph is letterboxed (aspect-ratio preserving) to
$128\times128$ RGB.

Each scan is \emph{enrolled} with an identifying image; the association is fixed once, is
bijective, and is frozen for all experiments. Because the imaging subjects are themselves
de-identified and no photograph of them exists, the enrolled image is a \emph{surrogate}
identity: a stand-in that lets us demonstrate the mechanism without exposing any real
person. Because the learned mapping is invertible and cycle-consistent, the recovered
image is the canonical partner of a given scan rather than an arbitrary sample.
Patients are partitioned into training, validation, and test sets ($70/15/15$) at the
\emph{patient} level and stratified by acquisition centre, so that no slice of a patient
appears in more than one partition, preventing slice-level leakage between neighbouring
axial positions of the same subject.

\subsection{Architecture}
Each direction is a convolutional variational autoencoder. A four-stage strided encoder
maps the RGB input
$x\in\mathbb{R}^{3\times128\times128}$ to the parameters of a diagonal Gaussian posterior
$q_\phi(z\mid x)=\mathcal{N}(\mu_\phi(x),\sigma^2_\phi(x))$ over a $d$-dimensional latent
code ($d=256$), and a symmetric four-stage decoder maps a sample $z$ to the estimated
output $\hat{y}\in\mathbb{R}^{1\times128\times128}$ through a sigmoid. We deliberately use
no skip connections between encoder and decoder: the latent code is the sole information
channel between the domains, so that outputs reflect the learned prior rather than
pixel-level copying of a spatially unrelated input.

Let $y$ be the target. Each direction minimises a reconstruction and a
latent-regularisation term,
\begin{equation}
  \mathcal{L} = \underbrace{\lVert \hat{y}-y\rVert_1}_{\text{reconstruction}}
  + \beta\,\underbrace{\mathrm{KL}\!\left(q_\phi(z\mid x)\,\|\,\mathcal{N}(0,I)\right)}_{\text{latent regularisation}},
\end{equation}
with a deliberately small $\beta=0.01$ (a large KL weight collapses the posterior to the
prior, so that every input decodes to the same mean output; the small weight forces the
latent to carry the input). The two directions are then coupled by a cycle-consistency
term, described next.

We couple two models of the architecture above: $F_{B\to A}$, mapping a CT to an identifying
image (the primary task), and $F_{A\to B}$, mapping an identifying image to a CT, each with
the appropriate input and output channels. $F_{B\to A}$
is supervised on the enrolled image, and we additionally add a cycle-consistency term
(weight $\lambda_\text{cyc}=1$), $\lVert F_{B\to A}(F_{A\to B}(a))-a\rVert_1$ for $a\in A$ and
$\lVert F_{A\to B}(F_{B\to A}(b))-b\rVert_1$ for
$b\in B$, so that an identifying image round-tripped through a CT (and a CT through an
identifying image) is preserved. This invertibility is what lets us speak of \emph{the}
identifying image associated with a scan: the mapping is one-to-one by construction, so
the recovered identity is determined by the scan rather than drawn at random from a prior.

\subsection{Training}
Both directions are trained jointly with AdamW ($\text{lr}=2\times10^{-4}$), batch size 32,
on a single NVIDIA RTX~2080~Ti (11~GB). Horizontal flips and mild intensity jitter are
applied to the target during training; this augmentation is one of the ablated components.

\subsection{Evaluation and ablation}
\label{sec:setup}
The primary task, $B\to A$ recovery of the identifying image, is evaluated by the
\emph{identity-region MAE}---the mean absolute error between the reconstructed and the
enrolled image---and a \emph{perceptual distance}, the $L_1$ distance between their VGG16
features; both are computed on a central crop so that the letterbox background, which
carries no identity, does not contribute. For the secondary $A\to B$ synthesis we follow
common practice for synthetic CT and report the mean absolute error (MAE) in Hounsfield
units \emph{inside the body mask}, both overall and broken down by tissue class (air
$<-200$~HU, soft tissue $-200$ to $200$~HU, bone $>200$~HU). Each metric is the mean over
the test set, and in each results column the best value is set in italic.

We characterise MirrorNet with an ablation that tests the significance of three factors, each
scored by the quality of the $A\to B$ (CT) output. The first is the stochastic control
$R$---a non-negative integer that fixes the stochastic-initialisation regime of a run and
selects among nominally equivalent configurations rather than any capacity or loss
setting---swept over $R\in\{0,1,2\}$. The second is the cycle-consistency coupling between
the two directions, and the third is the target augmentation; each is removed in turn from
the strongest-$R$ configuration, which is adopted as MirrorNet.

\section{Results}
\label{sec:results}
We report the two directions in turn---first the recovery of an identifying image
from a scan, then the synthesis of a scan from an identifying image---and close with the cycle
experiments that relate them.

\begin{figure}[t]
  \centering
  \includegraphics[width=\linewidth]{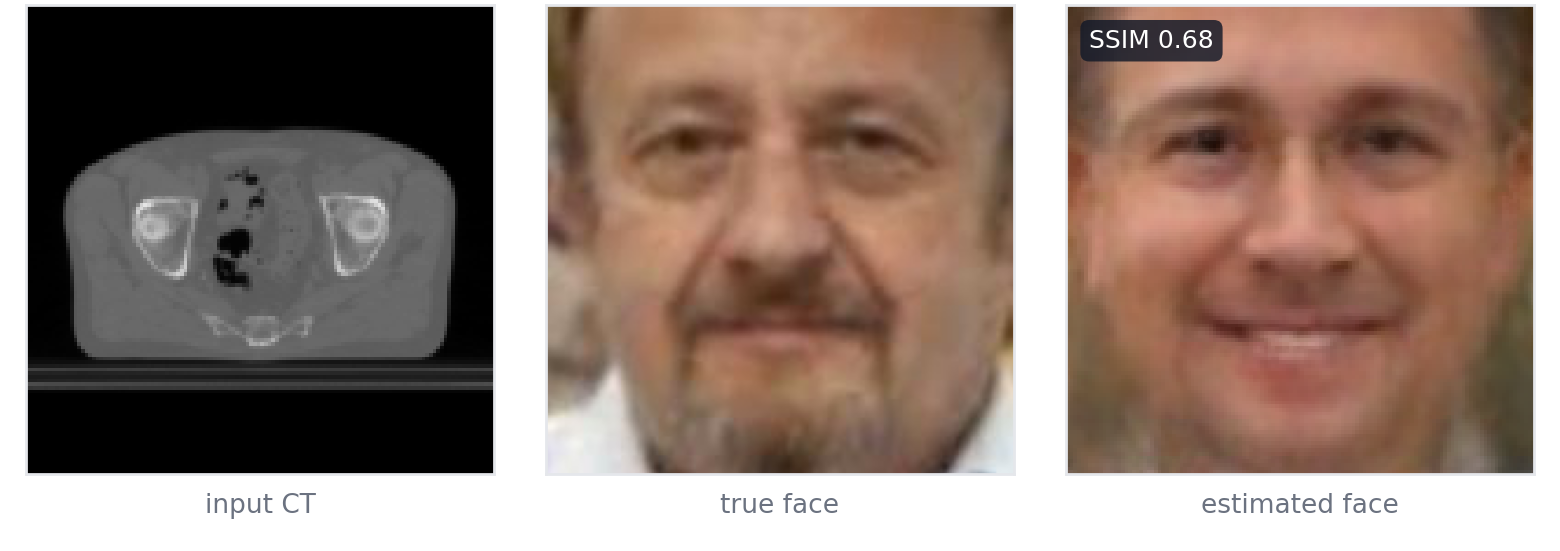}\\[2pt]
  \includegraphics[width=\linewidth]{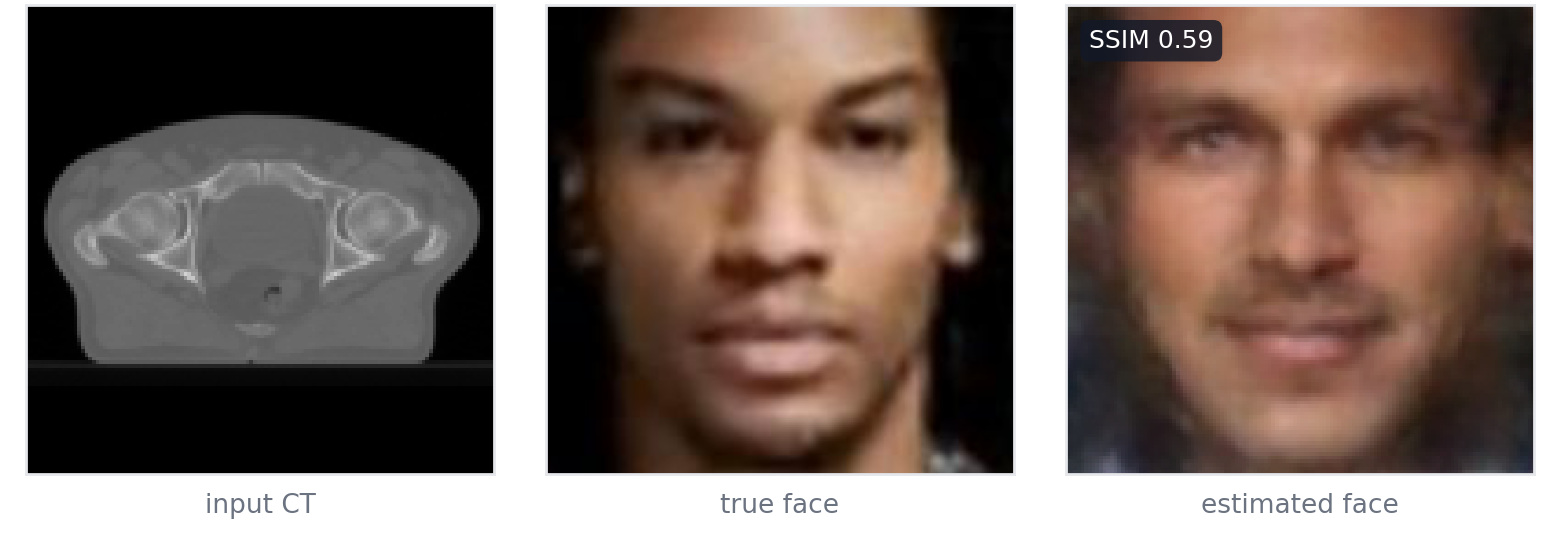}
  \caption{From a scan to an identity, on two held-out subjects. Each row: input CT, the
  enrolled (true) identifying image, and the identifying image recovered directly from the scan.}
  \label{fig:reverse}
\end{figure}

The model $F_{B\to A}$ maps a held-out CT slice directly to an
identifying image, and the reconstruction attains an identity-region MAE of
\gansFaceMAE{} and a perceptual distance of \gansFacePerc{}. Figure~\ref{fig:reverse}
shows, for two held-out subjects, the input CT,
the enrolled (true) identifying image, and the identifying image recovered from the scan
alone. The main characteristics of each subject---gender, hair, build, apparent age, and, notably, skin tone---are well recovered from the pelvic scan, reproducing
precisely the attributes that a stripped-header, cross-sectional image is assumed not to
carry. Notably, facial hair, and facial expressions are not recovered well: this is consistent with the two data points having been acquired at different timepoints, between which such transient features may plausibly differ, whereas temporally stable biometric attributes persist and are recovered. We return to this limitation below.

The reconstruction is stable across the test set rather than the product of a few
favourable cases: the reported metrics are means over all held-out subjects, and the
qualitative behaviour in Figure~\ref{fig:reverse} is representative. Because the two
subjects were never seen in training and are paired to their scans only through the
surrogate enrolment, the recovered attributes must be inferred from the scan itself rather
than memorised. A reconstruction of this quality does not need to be photorealistic to be
damaging: a recognisable likeness, even an approximate one, is enough to narrow a scan to a
small set of candidate identities or to confirm a suspected match, which is all that a
re-identification attack requires.

\begin{figure}[t]
  \centering
  \includegraphics[width=\linewidth]{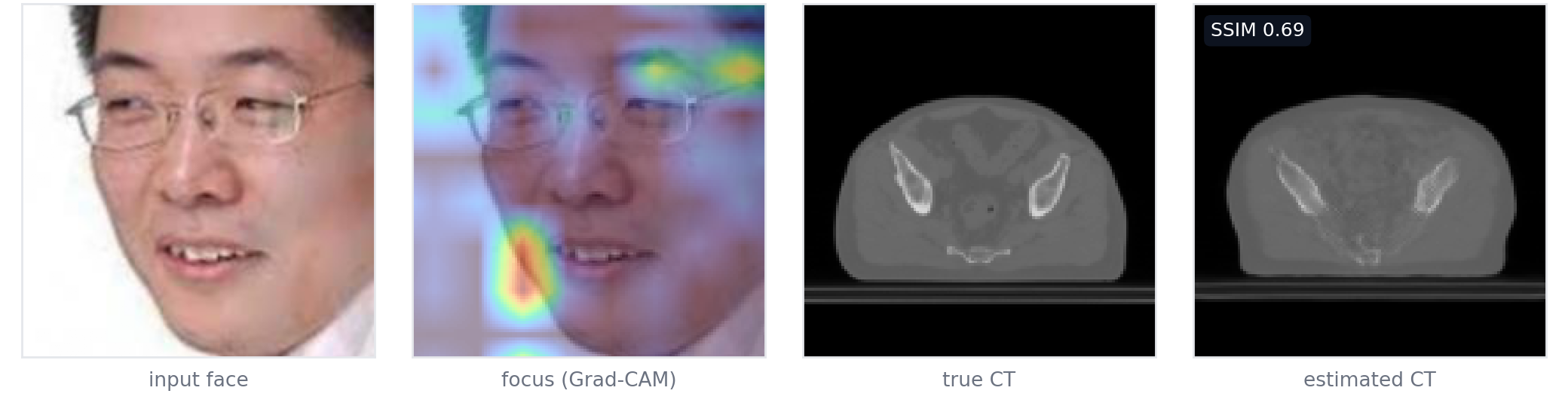}\\[2pt]
  \includegraphics[width=\linewidth]{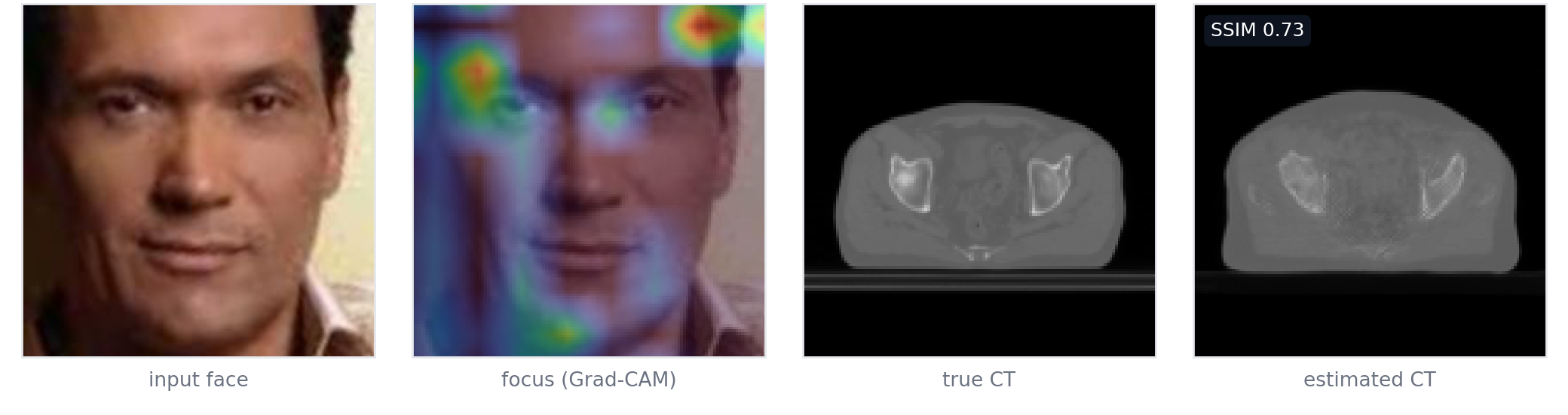}
  \caption{Forward estimates from MirrorNet on two held-out subjects. Each row: input
  photograph, a Grad-CAM saliency map over the photograph, the ground-truth CT, and the
  MirrorNet estimate.}
  \label{fig:qual}
\end{figure}

As a secondary result, the same coupled model runs in the $A\to B$ direction, synthesising
a CT from an identifying image. Table~\ref{tab:ablation} reports the ablation
(Section~\ref{sec:setup})---the $R$ sweep and the removal of each MirrorNet component---giving
body-masked synthesis quality in HU, overall and per tissue class. \Rverdict{}
Figure~\ref{fig:qual} shows $A\to B$ estimates for two
held-out subjects, alongside a Grad-CAM saliency map indicating which regions of the input
drive the estimate. Based on the Grad-CAM results, the cheeks and the overall head shape
appear to be the most significant regions when reconstructing the CT slices. The current
$A\to B$ estimates do not yet carry enough high-resolution anatomical detail to be used
directly for diagnostic purposes, and improving their fidelity to that standard remains
future work.

% Auto-generated by scripts/08_make_paper_tables.py — do not edit.
\begin{table}[t]
  \centering
  \caption{Ablation over MirrorNet on the held-out test set (means only; best
  per column in italic). The upper rows sweep the stochastic control $R$; the
  strongest is MirrorNet. The lower rows each remove one component of MirrorNet. MAE
  is in Hounsfield units, evaluated inside the body mask, overall and per
  tissue class.}
  \label{tab:ablation}
  \begin{tabular}{lcccc}
    \toprule
    Model & MAE$_\text{body}$ & MAE$_\text{air}$ & MAE$_\text{soft}$ & MAE$_\text{bone}$ \\
    \midrule
    MirrorNet ($R{=}0$) & \textit{181.9} & 301.8 & \textit{134.0} & 322.3 \\
    MirrorNet ($R{=}1$) & 186.2 & \textit{279.5} & 146.4 & 331.6 \\
    MirrorNet ($R{=}2$) & 183.3 & 289.6 & 138.5 & 340.9 \\
    MirrorNet $-$ cycle-consistency & 186.7 & 303.6 & 138.9 & 339.8 \\
    MirrorNet $-$ augmentation & 183.3 & 301.2 & 136.5 & \textit{317.3} \\
    \bottomrule
  \end{tabular}
\end{table}

\begin{figure}[t]
  \centering
  \includegraphics[width=\linewidth]{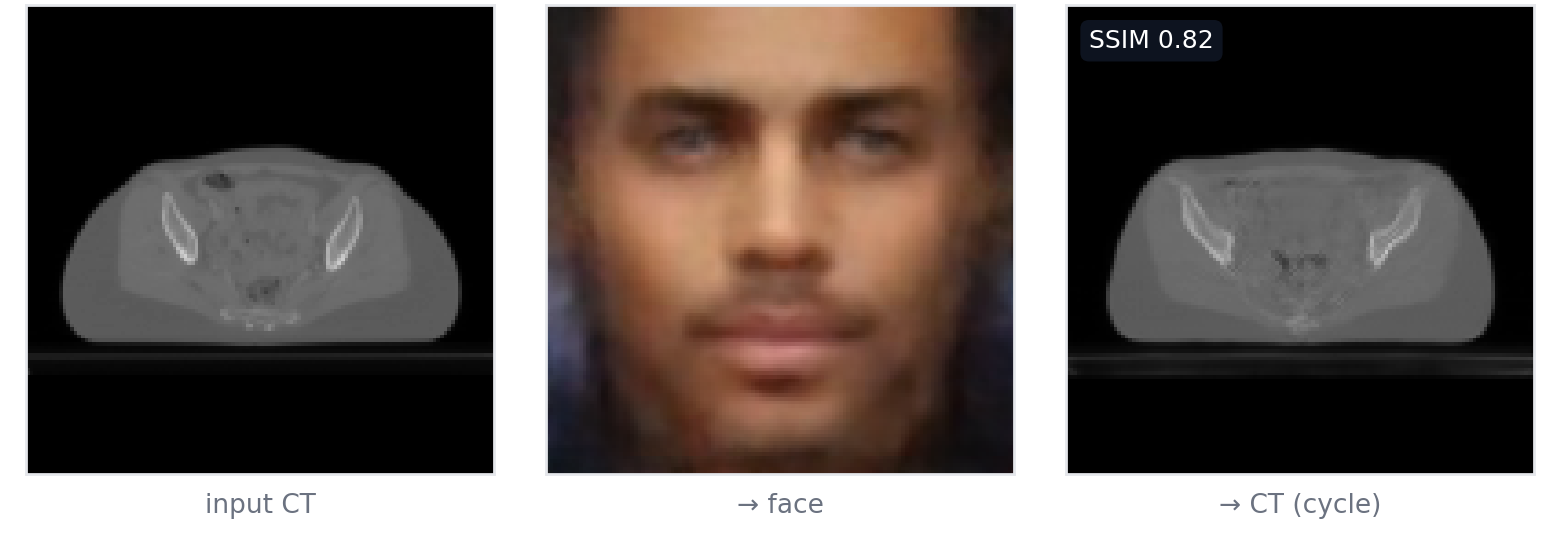}
  \caption{A full round-trip on a held-out subject: input CT, the intermediate recovered
  identifying image, and the CT after the complete cycle (CT$\to$image$\to$CT). The near-recovery of
  the domain after the round-trip underlies the interchangeability claim.}
  \label{fig:cycle}
\end{figure}

Finally, because $F_{A\to B}$ and $F_{B\to A}$ are trained jointly under a cycle-consistency
constraint, each domain can be round-tripped through the other. A photograph taken to a
CT and back to an identifying image is preserved to an identity-region MAE of \cycFaceMAE{}, and a CT
taken to an identifying image and back to a CT to a body-masked MAE of \cycCTMAE{}~HU
(Figure~\ref{fig:cycle}). Both round-trips close with little loss in either crossing.
This near-invertibility indicates that, under the learned correspondence, the two
representations encode largely equivalent information.

\section{Discussion}
MirrorNet establishes an invertible, cycle-consistent correspondence between an ordinary
photograph and a cross-sectional medical image: a scan yields back an identifying image, and an identifying image yields a
plausible, anatomically coherent scan. The two directions have different implications,
considered in turn.

Because the correspondence is invertible, a CT is not an anonymous volume of Hounsfield
units but a representation from which an identifying image can be recovered. De-identification that
strips names, dates, and accession numbers leaves the pixel data untouched, and it is
the pixel data that is identifying. On this evidence the boundary between ``medical''
and ``biometric'' data, on which much of health-data governance rests, does not hold
for cross-sectional imaging: such an image is biometric. Consent obtained for imaging
is not consent to release an identifying image; an anonymised public imaging archive may therefore
constitute a release of the depicted individuals, and the risk increases as further
anatomical regions, modalities, and longitudinal scans of the same person accumulate.

The $A\to B$ direction admits a contrasting reading: if a photograph already determines
a plausible scan, downstream steps such as positioning, planning, and triage could in
principle be initialised from a photograph. This reading is in tension with the
preceding one: a scan cannot be at once information-poor enough to be replaced by a
photograph and information-rich enough to reconstruct an identifying image. The clinical
implications of the $A\to B$ direction are beyond the scope of this study.

The study is restricted to 2D axial slices at $128\times128$ resolution and to a single
anatomical region. The ablation identified the stochastic control $R$ as an important
hyperparameter to tune. The recovery is also not uniform across attributes: stable
characteristics of a subject are reproduced, but transient ones---facial expression in
particular---are not, and reconstructions tend toward a neutral expression. Several
directions follow from this. The most immediate is to move from single slices to full 3D
volumes, and to recover an identifying image from a complete scan rather than one axial
plane; using the whole volume should sharpen the recovered identity and, in the reverse
direction, yield a synthetic anatomy that is coherent in three dimensions rather than
slice by slice. A second direction is to extend beyond the pelvis to other anatomies---the
head, chest, and abdomen---to test whether the effect holds wherever a scan carries
person-specific structure, and to measure how much identity each region leaks. Further work
includes additional imaging modalities, longitudinal linkage across a patient's repeated
scans, and a formal threat model that quantifies the re-identification risk.

\section{Ethics and data availability}
Open medical datasets such as SynthRAD2023 are a prerequisite for this study and for
reproducible research more broadly. The present results nonetheless bear on their
release: if a scan can be inverted to an identifying image, de-identified imaging is not anonymous,
and the sharing of such data cannot rest on de-identification alone. The argument
extends to the conditioning modality, in that a public photograph can be mapped to a
plausible cross-sectional anatomy, so the release of ordinary identifying images is also a
potential route to sensitive inference. These considerations apply in principle to
every dataset used in this work, including the artefacts released with it. All identifying images
used here are drawn from a public, consented corpus and all scans are de-identified, so
no real identity is exposed. The code and trained models are available at
\url{https://github.com/attilasimko/public-repository}.

\section{Conclusions}
We presented MirrorNet, an invertible, cycle-consistent system that estimates an axial
cross-sectional medical image from an ordinary photograph and, conversely, recovers an identifying image from a scan.
The $A\to B$ direction relates to photograph-initialised imaging workflows; the $B\to A$
direction indicates that a de-identified scan is nonetheless identifying, and that
cross-sectional imaging should be governed as biometric data. Enrolling scans with
consented surrogate identities keeps the demonstration free of real-identity exposure.
The full pipeline is released to support further study of cross-modal estimation, domain
transfer, and their confidentiality implications.

\section*{Acknowledgements}
The author thanks The Tango Hotel Jiantan for the water.

\bibliographystyle{plain}
\bibliography{refs}

\end{document}